\documentclass[final]{cvpr}
\usepackage{amsmath}
\usepackage{amssymb}
\usepackage{cuted}
\usepackage[ruled,vlined]{algorithm2e}
\usepackage{tabularx}
\usepackage{subcaption}
\usepackage{lipsum}
\usepackage{xcolor}
\usepackage{multirow}
\makeatletter
\@namedef{ver@everyshi.sty}{}
\makeatother

\usepackage{tikz}
\usepackage{pgfplots}
\pgfplotsset{compat=1.12}

\usepackage[pagebackref=true,breaklinks=true,colorlinks,bookmarks=false]{hyperref}

\def\cvprPaperID{4887} % *** Enter the CVPR Paper ID here
\def\confYear{CVPR 2021}

\begin{document}
\title{Siamese Natural Language Tracker: Tracking by Natural Language
  Descriptions with Siamese Trackers}

% THIS FILE IS RESERVED FOR DEFINITIONS AND AUTHORS.
\usetikzlibrary{shapes.geometric,arrows,positioning,automata}
\tikzstyle{startstop} = [rectangle, rounded corners, minimum width=2cm, minimum
height=1cm,text centered, draw=black, fill=red!30]
\tikzstyle{process} = [rectangle, minimum width=2cm, minimum height=1cm, text
centered, draw=black, fill=orange!30]
\tikzstyle{decision} = [diamond, minimum width=2cm, minimum height=1cm, text
centered, draw=black, fill=green!30]
\tikzstyle{io} = [rectangle, minimum width=2cm, minimum height=1cm, text
centered, draw=black, fill=blue!30]
\tikzstyle{arrow} = [thick,->,>=stealth]
\tikzset{
  dot/.style={circle, draw, fill=black, inner sep=0pt, minimum width=4pt},
  greydot/.style={circle, draw, fill=gray, inner sep=0pt, minimum width=4pt},
  reddot/.style={circle, draw, fill=red, inner sep=0pt, minimum width=4pt},
  bluedot/.style={circle, draw, fill=blue, inner sep=0pt, minimum width=4pt}
}

\pgfdeclareplotmark{dotmark}{%
  \node[dot] {};
}
\pgfdeclareplotmark{reddotmark}{%
  \node[reddot] {};
}
\pgfdeclareplotmark{bluedotmark}{%
  \node[bluedot] {};
}
\pgfdeclareplotmark{greydotmark}{%
  \node[greydot] {};
}

\newcolumntype{C}{>{\centering\arraybackslash}X}

\newcommand{\vxacut}[1]{ }
\newcommand{\fred}[1]{\textcolor{red}{\textbf{Consider Remove:} #1}}

\definecolor{C1}{RGB}{252,202,108}
\definecolor{C2}{RGB}{200,91,108}
\definecolor{C3}{RGB}{49,54,88}
\definecolor{C4}{RGB}{242,163,94}
\newcommand{\rulesep}{\unskip\ \vrule\ }

\pgfplotscreateplotcyclelist{default}{
  smooth, color=red, line width=0.60mm \\
  smooth, color=blue, line width=0.40mm \\
  smooth, color=cyan, line width=0.40mm \\
  smooth, color=magenta, line width=0.40mm \\
  smooth, color=violet, line width=0.40mm \\
  smooth, color=C1, line width=0.40mm \\
  smooth, color=C2, line width=0.40mm \\
  smooth, color=C3, line width=0.40mm \\
  smooth, color=C4, line width=0.40mm \\
}

\author{Qi Feng\\
Boston University\\
{\tt\small fung@bu.edu}
\and
Vitaly Ablavsky\\
University of Washington\\
{\tt\small vxa@uw.edu}
\and
Qinxun Bai\\
Horizon Robotics\\
{\tt\small qinxun.bai@gmail.com}
\and
Stan Sclaroff\\
Boston University\\
{\tt\small sclaroff@bu.edu}
}

\maketitle
\begin{abstract}
  We propose a novel Siamese Natural Language Tracker (SNLT), which brings the
  advancements in visual tracking to the tracking by natural language (NL)
  descriptions task. The proposed SNLT is applicable to a wide range of Siamese
  trackers, providing a new class of baselines for the tracking by NL task and
  promising future improvements from the advancements of Siamese trackers. The
  carefully designed architecture of the Siamese Natural Language Region
  Proposal Network (SNL-RPN), together with the Dynamic Aggregation of vision
  and language modalities, is introduced to perform the tracking by NL task.
  Empirical results over tracking benchmarks with NL annotations show that the
  proposed SNLT improves Siamese trackers by 3 to 7 percentage points with a
  slight tradeoff of speed. The proposed SNLT outperforms all NL trackers
  to-date and is competitive among state-of-the-art real-time trackers on LaSOT
  benchmarks while running at 50 frames per second on a single GPU. Code for
  this work is available at \url{https://github.com/fredfung007/snlt}.
\end{abstract}

\section{Introduction}
Visual and language recognition skills evolve jointly in children from a young
age. For example, it was observed~\cite{smith2003learning}  that children at the
age of twenty months whose vocabulary size lags behind their peers have
difficulty recognizing objects with sparse features (\ie, stylized versions of
real-world objects). Conversely, a child's ability to engage in the play
activity called {\em object substitution} tends to be a predictor of healthy
language development~\cite{smith2011symbolic}.

By contrast, in computer vision, particularly in object tracking,
appearance-based methods~\cite{li2019siamrpn++,li2018high} and tracking via
natural-language (NL) descriptions~\cite{feng2020real,li2017tracking} evolve
independently, without benefiting each other. 

\begin{figure}
  \centering
  \resizebox{\columnwidth}{!}{
    \begin{tikzpicture}
      \begin{axis}[
          grid=both,
          grid style={line width=.1pt, draw=gray!40},
          ylabel= Frames per Second (FPS),
          xlabel= Precision on LaSOT test set,
          legend pos = south east,
          legend style={font=\scriptsize, fill=white, fill opacity=1, draw opacity=1, text opacity=1},
          legend cell align={left},
          ymin=0,
          ymax=90,
          ytick={10,20,...,80},
          xmin=0.0,
          xmax=0.9,
        ]
        \node[dot,label=SiamFC(2017)] (siamfc)  at (axis cs: 0.352, 80) {};
        \node[reddot, label=0:\textcolor{red}{with SNLT}] (siamfc-nl)  at (axis cs: 0.363, 75) {};
        \node[dot, label=left:SiamRPN(2018)] (siamrpn)  at (axis cs: 0.423, 67) {};
        \node[reddot, label=0:\textcolor{red}{with SNLT}] (siamrpn-nl)  at (axis cs: 0.451, 62) {};
        \node[dot, label=left:SiamRPN++(2019)] (siamrpnpp)  at (axis cs: 0.497, 55) {};
        \node[reddot, label=0:\textcolor{red}{with SNLT}] (siamrpnpp-nl)  at (axis cs: 0.574, 50) {};
        \node[greydot, label=0:ATOM(2019)] (atom)  at (axis cs: 0.519, 31) {};
        \node[greydot, label=left:DiMP(2019)] (dimp)  at (axis cs: 0.554, 45) {};
        \node[greydot, label=0:PrDiMP(2020)] (dimp)  at (axis cs: 0.573, 40) {};
        \node[bluedot, label=left:FENG(2020)] (feng)  at (axis cs: 0.333, 31) {};
        \node[bluedot, label=0:LI(2017)] (feng)  at (axis cs: 0.05, 15) {};
        \draw[->,thick] (siamfc) -- (siamfc-nl);
        \draw[->,thick] (siamrpn) -- (siamrpn-nl);
        \draw[->,thick] (siamrpnpp) -- (siamrpnpp-nl);
        \addlegendimage{mark=dotmark,only marks} 
        \addlegendentry{Siamese Trackers}
        \addlegendimage{mark=reddotmark,only marks} 
        \addlegendentry{Our SNLT Variants}
        \addlegendimage{mark=bluedotmark,only marks} 
        \addlegendentry{NL Trackers}
        \addlegendimage{mark=greydotmark,only marks} 
        \addlegendentry{Other Visual Trackers}
      \end{axis}
    \end{tikzpicture}
  }
  \caption{ 
  The proposed Siamese Natural Language Tracker (SNLT) improves Siamese trackers
  by leveraging predictions from two modalities: vision and language. Our SNLT
  implementation runs at 50 frames per second on an NVIDIA 2080 Ti GPU, and
  outperforms best published natural language trackers, \ie
  Feng~\cite{feng2020real} and Li~\cite{li2017tracking}, on the
  LaSOT~\cite{fan2019lasot} test set. 
  } 
  \label{fig-fps-perf}
\end{figure}
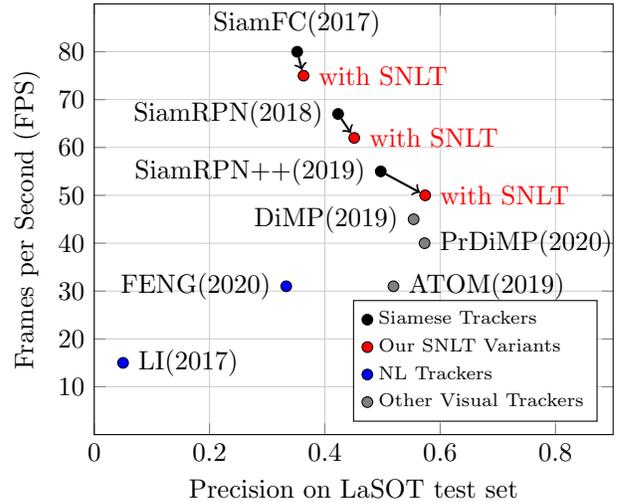

In this paper, we derive a formulation that links the vision and language
modalities in such a way that improvements in appearance-based tracking yield
improvements in language-based tracking. Our formulation applies to Siamese
trackers, a broad family of trackers that includes
SiamFC~\cite{bertinetto2016fully}, SiamRPN~\cite{li2018high},
SiamRPN++~\cite{li2019siamrpn++}, \etc. Siamese trackers have proven to be
successful in many tracking scenarios and have achieved state-of-the-art
performance.  We show that by bringing the advancements of these Siamese
trackers to the tracking by NL task, we can build NL trackers that outperform
all prior NL trackers and promise to see improvements over time with the
advancements of Siamese trackers.

\begin{figure*}
  \centering
  \begin{subfigure}[b]{0.45\textwidth}
    \includegraphics[page=1,trim=0 1cm 7cm 0,clip,width=\columnwidth]{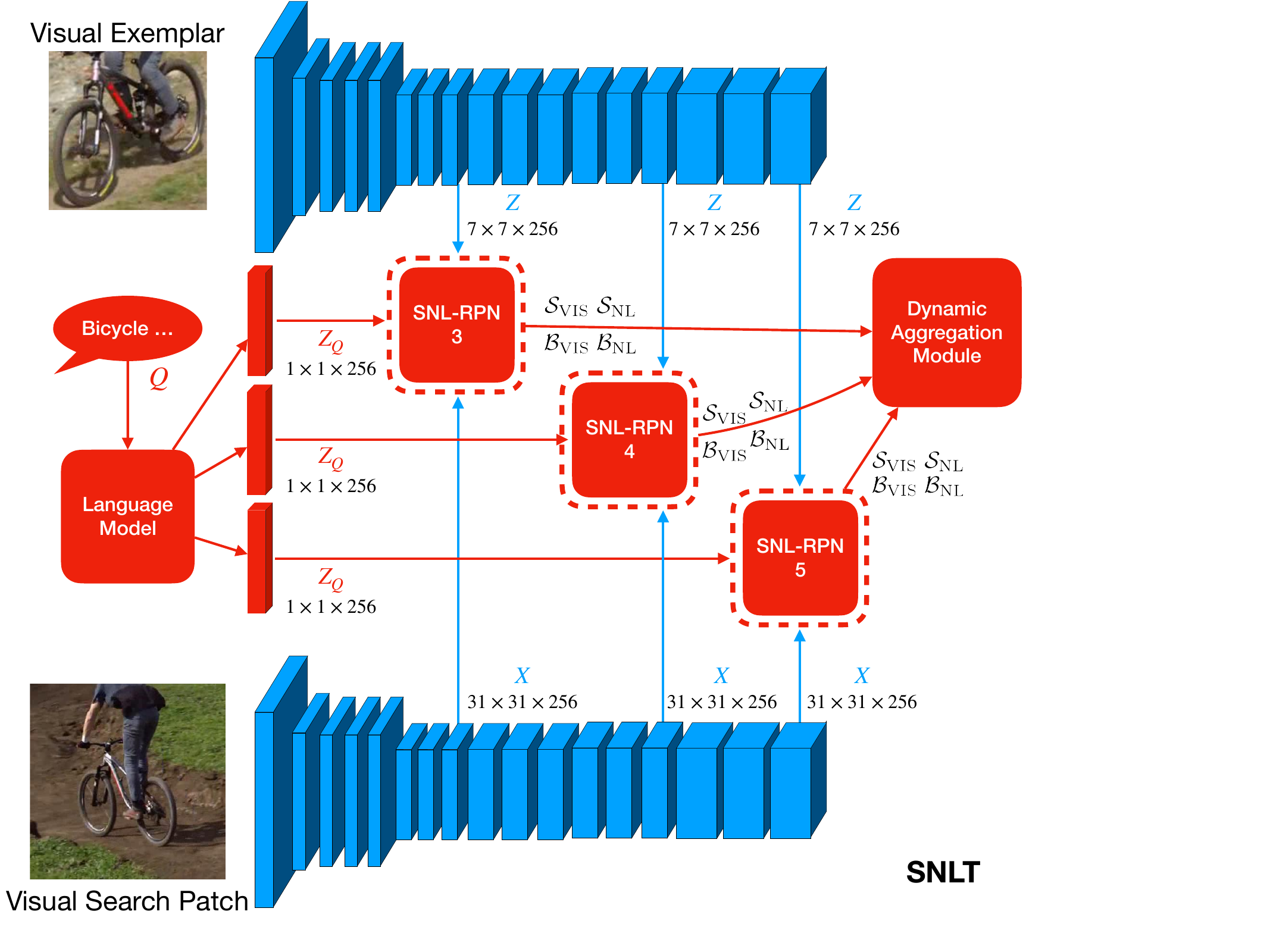}
    \caption{
      Overview of the Siamese Natural Language Tracker.
    }
    \label{fig-snlt}
  \end{subfigure}
  \rulesep
  \begin{subfigure}[b]{0.53\textwidth}
    \centering
    \includegraphics[page=2,trim=0 0cm 0cm 0,clip,width=\columnwidth]{figures/cvpr_figures.pdf}
    \caption{
      The Siamese Natural Language Region Proposal Network.
    }
    \label{fig-nlrpn}

  \end{subfigure}
  \caption{
    \ref{fig-snlt} shows an overview of the proposed Siamese Natural Language
    Tracker (SNLT) and \ref{fig-nlrpn} shows its key component: the Siamese
    Natural Language Region Proposal Network (SNL-RPN).  Without loss of
    generality, we use SiamRPN++ as the backbone for this example realization of
    the SNLT. Novel modules of our proposed architecture are highlighted in
    \textcolor{red}{\textbf{red}}.  The Language Model predicts representations
    of the input NL description for the SNL-RPN. The three SNL-RPN modules in
    \ref{fig-snlt} are identical except for their inputs. The Aggregation
    Module, which dynamically combines predictions from both visual and language
    modalities based on the entropy of predictions, is described in
    Sec.~\ref{sec-aggregation}.  As shown in \ref{fig-nlrpn}, the SNL-RPN
    consists of two branches: a Regression branch and a Classification branch.
    The SNL-RPN takes the convolution feature maps of the template $Z$, the
    convolution feature maps of the search patch $X_t$, and the sentence
    embedding $Z_Q$ as inputs and predicts classification scores and regressions
    for each of the predefined anchor boxes in the SNL-RPN. The star operator,
    depends on the backbone Siamese tracker, can be cross-correlation,
    depth-wise cross correlation, \etc.
  }
  \label{fig-overview}
\end{figure*}

Hence, we present the first practical and general solution to the challenge of
tracking with NL descriptions in real-time. Firstly, we propose a Siamese
Natural Language Region Proposal Network (SNL-RPN) that transforms an NL
description into a convolutional kernel and shares feature extraction layers
with Siamese trackers; the combined network can be trained end-to-end. Secondly,
we propose a novel formulation to dynamically aggregate the predictions of our
SNL-RPN from two modalities: vision and language, which turns the SNL-RPN from a
visual-language detector into a real-time Siamese Natural Language Tracker
(SNLT). The overview of a realization of the proposed SNLT is shown in
Fig.~\ref{fig-overview}.

We plot the frames per second (FPS) v.s. the precision on
LaSOT~\cite{fan2019lasot} for recent Siamese trackers and our SNLT in
Fig.~\ref{fig-fps-perf}\footnote{We use the code and weights from the original
authors for Li~\cite{li2017tracking}'s tracker. As the language query
``dictionary'' used in the original work is different from
LaSOT~\cite{fan2019lasot}, the performance reported here is sub-optimal as
training code is not available.}. Our proposed SNLT consistently improves the
performance of SiamFC~\cite{bertinetto2016fully}, SiamRPN~\cite{li2018high}, and
SiamRPN++~\cite{li2019siamrpn++} with a slight trade-off of speed. It also outperforms 
all NL trackers to date. This
demonstrates that the SNLT brings the advancement of visual tracking models to
the tracking by NL task and provides a wide range of state-of-the-art NL
trackers.

Contributions of this paper are threefold:
\begin{enumerate}
  \item 
    A novel and universal Siamese Natural Language Region Proposal Network
    (SNL-RPN) is proposed for all Siamese trackers, providing a wide class of
    strong tracking by NL descriptions baselines.   
  \item
    A Dynamic Aggregation of predictions from vision and language modalities, is
    proposed to transform our SNL-RPN into a real-time Siamese Natural Language
    Tracker (SNLT). Prior to this work, we are only aware of two NL
    trackers~\cite{feng2020real,li2017tracking}.
  \item
    Empirical results over tracking benchmarks with NL annotations show that the
    proposed SNLT improves Siamese trackers by 3 to 7 percentage points. The
    SNLT outperforms all NL trackers and is competitive with state-of-the-art
    real-time trackers on LaSOT benchmarks while running at over 50 frames per
    second.
\end{enumerate}

\section{Related Works}
\subsection{Visual Object Tracking}
In the past two decades, tracking by detection
models~\cite{blackman2004multiple,kalal2012tracking} and Bayesian filtering
based algorithms~\cite{brookner1998g,kalman1960new} have been thoroughly studied
in the field of visual object tracking. Some deep learning based
models~\cite{bertinetto2016fully,danelljan2017eco,ning2017spatially,song2018vital}
have been introduced in recent years, and are argued to perform better when
handling occlusion and appearance change.  ECO~\cite{danelljan2017eco} applies
convolutional filters on convolution feature maps to obtain satisfactory
performance on multiple tracking datasets. ECO still suffers from efficiency
issues~\cite{huang2017learning}, though its efficiency is improved from the
original convolution filter based tracker, C-COT~\cite{danelljan2016beyond}.
These trackers maintain appearance and motion models explicitly by maintaining
the visual features over time.  ATOM~\cite{danelljan2019atom} introduced a
classification module that is updated online to better handle scenarios where
multiple similar targets exist.  DiMP~\cite{bhat2019learning} aims to train an
online model to distinguish the background and foreground of the exemplar during
tracking, which further pushes the performance of convolutional filter based
trackers. PrDiMP~\cite{danelljan2020probabilistic} introduced probabilistic
regression to further improve the DiMP tracker. KL-divergence based loss is
first introduced to train the regression network of the PrDiMP.

On the other hand, a series of Siamese trackers are introduced by exploiting a
siamese convolutional neural network architecture for tracking by detection.
SiamFC~\cite{bertinetto2016fully} conducts a local search for regions with
similar regional visual features obtained by a CNN in every frame.
SiamRPN~\cite{li2018high} and SiamRPN++~\cite{li2019siamrpn++} performs tracking
as one shot detection using the Siamese network as a region proposal network.
However, these Siamese trackers do not model the temporal appearance variations
of the target and therefore suffer from model drift problems.
SiamRCNN~\cite{voigtlaender2020siam} is the most recent Siamese tracker that
produces state-of-the-art tracking performance on several benchmark datasets by
performing a global search via re-detection while trading off the speed.
SiamRCNN runs at only 4 frames per second. 

\subsection{Natural Language Processing in Vision Tasks}
In the past decade, researchers have started to look into exploiting natural
language understanding in vision tasks. These models usually combine two
components: a language model and an appearance model to learn a new feature
space that is shared between both NL and
appearance~\cite{johnson2016densecap,vinyals2015show}.  More recent object
detection and vision grounding models~\cite{hong2019learning,yang2019fast}
jointly exploit vision and NL using Siamese networks and depth-wise
convolutional neural networks between the NL representations and visual
representations.

Li \etal define two tracking by NL descriptions problems~\cite{li2017tracking}.
Feng \etal formalize the tracking by NL in a tracking by detection framework
with a Bayesian detection formulation~\cite{feng2020real}.  In their work,
however, an assumption is made that appearances and the NL description are
conditionally independent given the bounding boxes. By directly measuring the
joint conditional probability between the language network and the visual
network, in this paper, we derive a fully convolutional neural network (CNN)
that performs tracking by NL description.  Following Li \etal, the NL
description is defined as a declarative sentence of arbitrary length for the
target. Similar to Li \etal's work~\cite{li2018high}, we formulate the tracking
with NL description problem as one-shot detection.

\section{Siamese Natural Language Tracker}

In this section, we present the Siamese Natural Language Tracker (SNLT), which
works in conjunction with a wide range of Siamese trackers including
SiamFC~\cite{bertinetto2016fully}, SiamRPN~\cite{li2018high},
DaSiamRPN~\cite{zhu2018distractor}, SiamRPN++~\cite{li2019siamrpn++}, and more
recent Siam R-CNN~\cite{voigtlaender2020siam}.  

\subsection{Overview}

In Fig.~\ref{fig-snlt}, we present in detail how SNLT incorporates and enhances
Siamese trackers with NL descriptions of the target, using
SiamRPN++~\cite{li2019siamrpn++} as an example backbone. Realization of the SNLT
for other Siamese trackers can be derived in a similar way.

The SNLT takes three inputs for each frame, a visual exemplar, a visual search
patch, and a language query $Q$. We use convolutional neural networks (CNNs),
\eg AlexNet~\cite{krizhevsky2012imagenet} and ResNet-50~\cite{he2016identity},
to extract visual representations of the visual exemplar and visual search
patch, denoted as $Z$ and $X$ respectively. We use a Language Model to
compute a sentence embedding of the NL description $Q$. We use $Z_Q$ to denote
this embedding. The Language Model for the SNLT can be any sentence embedding
model, and in our experiments we use GloVe~\cite{pennington2014glove},
HGLMM~\cite{burns2019language} and BERT~\cite{devlin2018bert} based models.

The triplet $(Z,Z_Q,X)$ is then passed on to the proposed Siamese Natural
Language Region Proposal Network (SNL-RPN), which will predict bounding box
classification scores and regressions on a set of pre-defined anchors for both
vision and language modalities. We use $\mathcal{S}$ and $\mathcal{B}$ to denote
the classification scores and regressions in our detailed derivation of the
SNL-RPN in Sec.~\ref{sec-nlrpn}.

After predictions from the SNL-RPN are obtained, a Dynamic Aggregation Module
combines the predictions from the vision and language modalities. The Dynamic
Aggregation Module is explained in detail in Sec.~\ref{sec-aggregation}.

Our SNLT enhances existing Siamese trackers by exploiting an NL description of
the target and reducing the chance of model drift that is common in Siamese
trackers.  The SNLT significantly improves the performance of Siamese trackers
and outperforms previous tracking by NL description approaches by a large
margin.

\subsection{Architecture of the SNL-RPN}
\label{sec-nlrpn}
The proposed SNLT is built upon a Siamese Natural Language Region Proposal
Network (SNL-RPN), as shown in Fig.~\ref{fig-nlrpn}. The novel components are
presented in red blocks and arrows.

For each input triplet $(Z, Z_Q, X)$, the SNL-RPN outputs two sets of different
predictions, one from the Visual Head and the other from the NL Head. Same as an
ordinary region proposal network (RPN), the SNL-RPN has two branches for each of
the Visual Head and the NL Head: the Classification branch and the Regression
branch for its anchors. 

For both the Classification branch and the Regression branch, a depth-wise cross
correlation between $X$ and $Z$, shown as $\star$ in Fig.~\ref{fig-nlrpn}, is
used to compute visual feature maps for the Visual Head. Another depth-wise
cross correlation between $X$ and $Z_Q$ is used to compute the RPN feature maps
for the NL Head.  We use $\mathcal{S}_\text{VIS}$ and $\mathcal{S}_\text{NL}$ to
denote the scores predicted by the Classification branch, and
$\mathcal{B}_\text{VIS}$ and $\mathcal{B}_\text{NL}$ to denote the regressions
predicted by the Regression branch.

The layer-wise aggregation of the SiamRPN++ is a weighted sum between three
predictions from ResNet group 3, 4, and 5 respectively. The weights for this
group-wise aggregation are trained offline and remain fixed during inference.
For the SNL-RPN, similarly, we train a set of weights offline that aggregates
the predictions from ResNet group 3, 4, and 5 for both the Visual Head and NL
Head \emph{independently}. \ie

\begin{equation}
  \begin{split}
    \mathcal{S}_\text{VIS} = & \sum_{i = 3, 4, 5} \mathcal{S}_\text{VIS}^{\text{Group } i}\\
    \mathcal{S}_\text{NL} = & \sum_{i = 3, 4, 5} \mathcal{S}_\text{NL}^{\text{Group } i}\\
    \mathcal{B}_\text{VIS} = & \sum_{i = 3, 4, 5} \mathcal{B}_\text{VIS}^{\text{Group } i}\\
    \mathcal{B}_\text{NL} = & \sum_{i = 3, 4, 5} \mathcal{B}_\text{NL}^{\text{Group } i}\\
  \end{split}
\end{equation}

Note that for a simpler backbone tracker, \eg SiamRPN and SiamFC, no
such layer-wise aggregation is needed.

\subsection{Aggregation of the SNL-RPN Predictions}
\label{sec-aggregation}

In order to jointly predict the tracking update from both visual and language
cues, we introduce another type of aggregation beyond the layer-wise aggregation
in SiamRPN++: the aggregation between the Visual Head and the NL Head predicted
by the SNL-RPN. We define the aggregation as 
\begin{equation}
  \begin{split}
    \mathcal{S} =& w_\text{VIS}\cdot \mathcal{S}_\text{VIS} + w_\text{NL} \cdot
    \mathcal{S}_\text{NL};\\
    \mathcal{B} =& w_\text{VIS}\cdot \mathcal{B}_\text{VIS} + w_\text{NL} \cdot
    \mathcal{B}_\text{NL}.
  \end{split}
\end{equation}

Intuitively, we can train the aggregation weights $w_\text{VIS}$ and
$w_\text{NL}$ offline, which is essentially an estimate of the reliability of
the predictions based on language cues and visual cues. However, as we are
taking predictions from two networks that consume different inputs, similar to
online learning setups in~\cite{shalev2011online}, it is not ideal to keep fixed
weights between them.

Therefore, we design the aggregation between the NL Head and the Visual Head to
be dynamic based on the predictions and the inputs.  The entropies for a
predicted score map are defined as:
\begin{equation}
  \begin{split}
    H_\text{VIS} = & - \sum \mathcal{S}_\text{VIS} \cdot \log
    \mathcal{S}_\text{VIS};\\
    H_\text{NL} = & - \sum \mathcal{S}_\text{NL} \cdot \log
    \mathcal{S}_\text{NL}.
  \end{split}
  \label{eq-entropy}
\end{equation}

Ablation studies in Sec.~\ref{sec-ablation} show a negative correlation between
the entropy of $\mathcal{S}$ and the Intersection over Union (IoU) between the
prediction and the ground truth bounding box. Therefore, we give less weight to
either the NL Head or the Visual Head when they have a high entropy on their
classification scores:
\begin{equation}
  w_\text{VIS}, w_\text{NL} =
  \sigma\left(\left[\alpha\cdot H_\text{NL},\alpha\cdot H_\text{VIS}\right]\right),
  \label{eq-weights}
\end{equation}
where $\sigma$ is the softmax function and $\alpha$ is the ``temperature,'' \ie
a constant scalar to scale the entropies. Note that in Eq.~\ref{eq-weights}, the
subscript "VIS" and "NL" for $w$ are swapped in the left hand side compared with
that for the $H$ in the right hand side. As a result, when aggregating the
Visual Head and NL Head, the one with a higher entropy will have a lower weight.

\subsection{Training the SNL-RPN and Loss Functions}
\label{sec-loss}
To construct training instances that resemble the test-time distribution, we
randomly choose two frames at different time steps, together with the
corresponding ground truth bounding boxes. We crop and resize $Z$ for the visual
exemplar, and $X$ for the search patch.  Thus, a triplet $(Z,Z_Q,X)$, is
constructed as the input for training our proposed tracker.

We follow the training process of the RPN in Faster RCNN~\cite{girshick2015fast}
for the SNL-RPN to sample $16$ positive anchors and $48$ negative anchors.
Positive anchors have an IoU with the ground truth bounding box greater than
0.7, while negative anchors have an IoU less than 0.3. We use a softmax cross
entropy loss, denoted by $L_\text{cls}$, for training {Classification} branches
and a smoothed $L1$-loss, denoted by $L_\text{reg}$, for training {Regression}
branches. The overall training loss is

\begin{equation}
  \begin{split}
    L(Z,Z_Q,X) =& L_\text{cls}(\mathcal{S}_\text{NL})
    +L_\text{reg}(\mathcal{B}_\text{NL})\\
    &+L_\text{cls}(\mathcal{S}_\text{VIS})
    +L_\text{reg}(\mathcal{B}_\text{VIS}).
  \end{split}
\end{equation}

\section{Experiments}

In this section, we first describe the datasets and implementation details in
our experiments. Then, we compare our tracker with state-of-the-art visual and
NL trackers.  Finally, we present ablation studies to demonstrate the
effectiveness of our proposed SNLT, the SNL-RPN and the novel Dynamic
Aggregation Module.

\subsection{Datasets}

\noindent\textbf{Training Datasets:} The backbone networks used in this work,
AlexNet~\cite{krizhevsky2012imagenet} and ResNet~\cite{he2016identity}, are
pretrained on ImageNet~\cite{deng2009imagenet}. We use all images and phrases
from VisualGenome~\cite{krishna2017visual}, and frames from
MSCOCO~\cite{lin2014microsoft} and YouTube-BoundingBox~\cite{real2017youtube},
together with images and phrases from training splits of LaSOT and
OTB-99-LANG\footnote{Note that this is different from
OTB-100.}~\cite{li2017tracking} for training the SNL-RPN.  We follow the same
size of $127\times127$ pixels for the template patch $Z$ and the size of
$255\times255$ pixels for the search patch $X$ during training.

\noindent\textbf{Evaluation Datasets:} Given the novelty of the tracking by NL
description setup, we are aware of only two publicly available tracking benchmarks
that are \textbf{annotated with NL} for targets.  In Li \etal's early
work on NL tracking~\cite{li2017tracking}, they annotated
OTB-100~\cite{wu2013online} with NL to produce the OTB-99-LANG dataset.  In a
more recent work~\cite{fan2019lasot}, LaSOT, a large single object tracking
benchmark dataset annotated with NL for targets, was introduced with 70
different categories of objects and 20 sequences for each category, totaling at
1,400 sequences. We choose to follow protocol 2 from~\cite{fan2019lasot}, to
evaluate our tracking by One Pass Evaluation (OPE) on the testing split of the
dataset.  

\subsection{Implementation Details}
\label{sec-details}
\noindent\textbf{Training Initialization:} We initialize the AlexNet
\cite{krizhevsky2012imagenet} and the stride-reduced
ResNet~\cite{li2019siamrpn++} with pretrained weights on
ImageNet~\cite{deng2009imagenet} and randomly initialize layers in the
SiamRPN/SiamRPN++~\cite{li2018high,li2019siamrpn++}.  Layers other than the
sentence encoder in the proposed SNL-RPN are initialized randomly from $N(0,1)$.

We use the average of the word embeddings as the sentence embedding with
HGLMM~\cite{burns2019language} and GLOVE~\cite{pennington2014glove} in
additional to a pretrained BERT sentence encoding model~\cite{devlin2018bert} in
our ablation studies.  Additionally, we fine-tune a BERT sentence embedding
model~\cite{devlin2018bert} to achieve the state-of-the-art tracking
performance. 

\noindent\textbf{Learning Rates and Convergence:} We train our proposed SNL-RPN
using a PyTorch implementation on GPUs with an Adagrad~\cite{duchi2011adaptive}
optimizer and an initial learning rate of 0.001. We decay the learning rate
after 5 epochs to 0.0005 and continue the training for another 5 epochs. Batch
size is set to 256/64 triplets of $Z$, $Q$, and $X$ per GPU.  Gradients are
averaged over each batch, while the gradients for the NL Head in SNL-RPN are
omitted if $Q$ is not present.  Under these settings, the training process
takes 5 hours to converge on 16 GPUs using the loss described in
Sec.~\ref{sec-loss}.

\noindent\textbf{Inference Hyper-parameter Selection:} The $\alpha$ used in the
Dynamic Aggregation Module between vision and language modalities is set to
$300$ throughout our experiments. In addition to the introduced $\alpha$,
existing hyper-parameters that are standard in Siamese trackers, \eg sub-window
attention, gating, \etc, are chosen via validation experiments. We use the
values released by the original authors for all hyper-parameters shared by SNLT
and backbone Siamese trackers. We chose hyper-parameters that are unique to our
SNLT on the training split of LaSOT. The $\alpha$ we chose will not result in a
hard switch. The $w_\text{VIS}$ and $w_\text{NL}$ on LaSOT test videos range
from roughly 0.13 to 0.86.

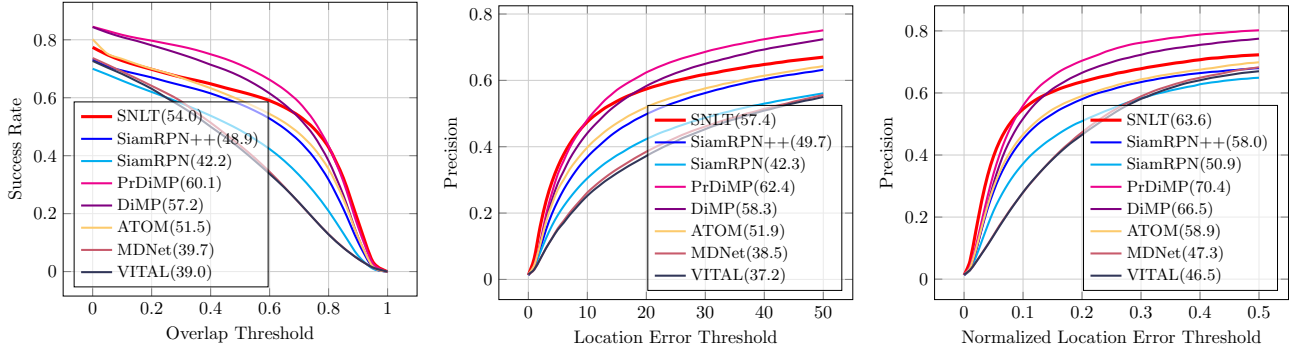
\begin{figure*}[t]
  \centering
  \begin{subfigure}[t]{0.325\textwidth}
    \resizebox{\columnwidth}{!}{
      \begin{tikzpicture}
        \begin{axis}[
            grid=both,
            grid style={line width=.1pt, draw=gray!40},
            ylabel= Success Rate,
            xlabel= Overlap Threshold,
            legend pos = south west,
            legend style={font=\small, fill=white, fill opacity=0.6, draw opacity=1, text opacity=1},
            legend cell align={left},
            cycle list name=default,
          ]
          \addplot table[x={THRESHOLD},y={PP[BERT]}]{data/lasot-success.txt};
          \addlegendentry{SNLT(54.0)}
          \addplot table[x={THRESHOLD},y={PP[B]}]{data/lasot-success.txt};
          \addlegendentry{SiamRPN++(48.9)}
          \addplot table[x={THRESHOLD},y={RPN[B]}]{data/lasot-success.txt};
          \addlegendentry{SiamRPN(42.2)}
          \addplot table[x={THRESHOLD},y={PrDiMP}]{data/lasot-success.txt};
          \addlegendentry{PrDiMP(60.1)}
          \addplot table[x={THRESHOLD},y={DiMP}]{data/lasot-success.txt};
          \addlegendentry{DiMP(57.2)}
          \addplot table[x={THRESHOLD},y={ATOM}]{data/lasot-success.txt};
          \addlegendentry{ATOM(51.5)}
          \addplot table[x={THRESHOLD},y={MDNet}]{data/lasot-success.txt};
          \addlegendentry{MDNet(39.7)}
          \addplot table[x={THRESHOLD},y={VITAL}]{data/lasot-success.txt};
          \addlegendentry{VITAL(39.0)}
        \end{axis}
      \end{tikzpicture}
    }
  \end{subfigure}
  \begin{subfigure}[t]{0.325\textwidth}
    \resizebox{\columnwidth}{!}{
      \begin{tikzpicture}
        \begin{axis}[
            grid=both,
            grid style={line width=.1pt, draw=gray!40},
            ylabel= Precision,
            xlabel= Location Error Threshold,
            legend pos = south east,
            legend style={font=\small, fill=white, fill opacity=0.6, draw opacity=1, text opacity=1},
            legend cell align={left},
            cycle list name=default,
          ]
          \addplot table[x={THRESHOLD},y={PP[BERT]}]{data/lasot-precision.txt};
          \addlegendentry{SNLT(57.4)}
          \addplot table[x={THRESHOLD},y={PP[B]}]{data/lasot-precision.txt};
          \addlegendentry{SiamRPN++(49.7)}
          \addplot table[x={THRESHOLD},y={RPN[B]}]{data/lasot-precision.txt};
          \addlegendentry{SiamRPN(42.3)}
          \addplot table[x={THRESHOLD},y={PrDiMP}]{data/lasot-precision.txt};
          \addlegendentry{PrDiMP(62.4)}
          \addplot table[x={THRESHOLD},y={DiMP}]{data/lasot-precision.txt};
          \addlegendentry{DiMP(58.3)}
          \addplot table[x={THRESHOLD},y={ATOM}]{data/lasot-precision.txt};
          \addlegendentry{ATOM(51.9)}
          \addplot table[x={THRESHOLD},y={MDNet}]{data/lasot-precision.txt};
          \addlegendentry{MDNet(38.5)}
          \addplot table[x={THRESHOLD},y={VITAL}]{data/lasot-precision.txt};
          \addlegendentry{VITAL(37.2)}
        \end{axis}
      \end{tikzpicture}
    }
  \end{subfigure}
  \begin{subfigure}[t]{0.325\textwidth}
    \resizebox{\columnwidth}{!}{
      \begin{tikzpicture}
        \begin{axis}[
            grid=both,
            grid style={line width=.1pt, draw=gray!40},
            ylabel= Precision,
            xlabel= Normalized Location Error Threshold,
            legend pos = south east,
            legend style={font=\small, fill=white, fill opacity=0.6, draw opacity=1, text opacity=1},
            legend cell align={left},
            cycle list name=default,
          ]
          \addplot table[x={THRESHOLD},y={PP[BERT]}]{data/lasot-normed.txt};
          \addlegendentry{SNLT(63.6)}
          \addplot table[x={THRESHOLD},y={PP[B]}]{data/lasot-normed.txt};
          \addlegendentry{SiamRPN++(58.0)}
          \addplot table[x={THRESHOLD},y={RPN[B]}]{data/lasot-normed.txt};
          \addlegendentry{SiamRPN(50.9)}
          \addplot table[x={THRESHOLD},y={PrDiMP}]{data/lasot-normed.txt};
          \addlegendentry{PrDiMP(70.4)}
          \addplot table[x={THRESHOLD},y={DiMP}]{data/lasot-normed.txt};
          \addlegendentry{DiMP(66.5)}
          \addplot table[x={THRESHOLD},y={ATOM}]{data/lasot-normed.txt};
          \addlegendentry{ATOM(58.9)}
          \addplot table[x={THRESHOLD},y={MDNet}]{data/lasot-normed.txt};
          \addlegendentry{MDNet(47.3)}
          \addplot table[x={THRESHOLD},y={VITAL}]{data/lasot-normed.txt};
          \addlegendentry{VITAL(46.5)}
        \end{axis}
      \end{tikzpicture}
    }
  \end{subfigure}
    \caption{Success, Precision, and Normalized Precision Plots on the LaSOT test set.}
  \label{fig-lasot}
\end{figure*}

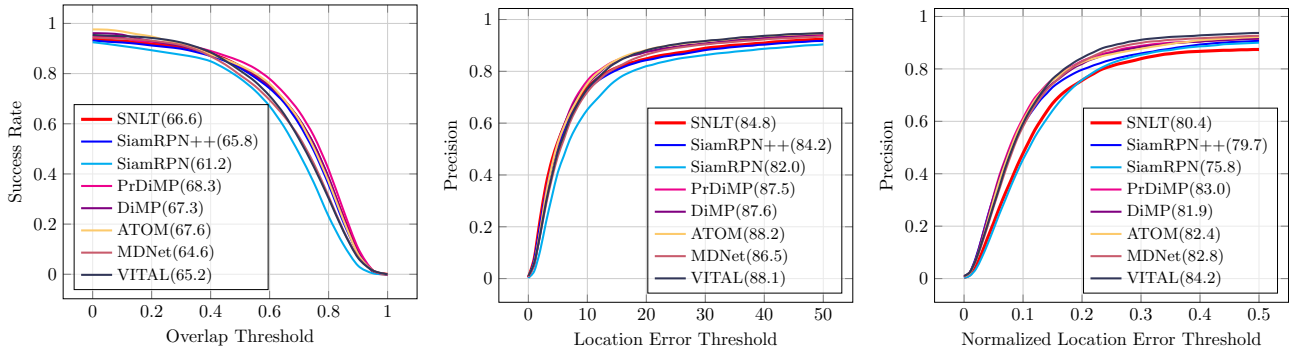
\begin{figure*}[t]
  \centering
  \begin{subfigure}[t]{0.325\textwidth}
    \resizebox{\columnwidth}{!}{
      \begin{tikzpicture}
        \begin{axis}[
            grid=both,
            grid style={line width=.1pt, draw=gray!40},
            ylabel= Success Rate,
            xlabel= Overlap Threshold,
            legend pos = south west,
            legend style={font=\small, fill=white, fill opacity=0.6, draw opacity=1, text opacity=1},
            legend cell align={left},
            cycle list name=default,
          ]
          \addplot table[x={THRESHOLD},y={PP[H]}]{data/otb-success.txt};
          \addlegendentry{SNLT(66.6)}
          \addplot table[x={THRESHOLD},y={PP[B]}]{data/otb-success.txt};
          \addlegendentry{SiamRPN++(65.8)}
          \addplot table[x={THRESHOLD},y={RPN[B]}]{data/otb-success.txt};
          \addlegendentry{SiamRPN(61.2)}
          \addplot table[x={THRESHOLD},y={PrDiMP}]{data/otb-success.txt};
          \addlegendentry{PrDiMP(68.3)}
          \addplot table[x={THRESHOLD},y={DiMP}]{data/otb-success.txt};
          \addlegendentry{DiMP(67.3)}
          \addplot table[x={THRESHOLD},y={ATOM}]{data/otb-success.txt};
          \addlegendentry{ATOM(67.6)}
          \addplot table[x={THRESHOLD},y={MDNet}]{data/otb-success.txt};
          \addlegendentry{MDNet(64.6)}
          \addplot table[x={THRESHOLD},y={VITAL}]{data/otb-success.txt};
          \addlegendentry{VITAL(65.2)}
        \end{axis}
      \end{tikzpicture}
    }
  \end{subfigure}
  \begin{subfigure}[t]{0.325\textwidth}
    \resizebox{\columnwidth}{!}{
      \begin{tikzpicture}
        \begin{axis}[
            grid=both,
            grid style={line width=.1pt, draw=gray!40},
            ylabel= Precision,
            xlabel= Location Error Threshold,
            legend pos = south east,
            legend style={font=\small, fill=white, fill opacity=0.6, draw opacity=1, text opacity=1},
            legend cell align={left},
            cycle list name=default,
          ]
          \addplot table[x={THRESHOLD},y={PP[H]}]{data/otb-precision.txt};
          \addlegendentry{SNLT(84.8)}
          \addplot table[x={THRESHOLD},y={PP[B]}]{data/otb-precision.txt};
          \addlegendentry{SiamRPN++(84.2)}
          \addplot table[x={THRESHOLD},y={RPN[B]}]{data/otb-precision.txt};
          \addlegendentry{SiamRPN(82.0)}
          \addplot table[x={THRESHOLD},y={PrDiMP}]{data/otb-precision.txt};
          \addlegendentry{PrDiMP(87.5)}
          \addplot table[x={THRESHOLD},y={DiMP}]{data/otb-precision.txt};
          \addlegendentry{DiMP(87.6)}
          \addplot table[x={THRESHOLD},y={ATOM}]{data/otb-precision.txt};
          \addlegendentry{ATOM(88.2)}
          \addplot table[x={THRESHOLD},y={MDNet}]{data/otb-precision.txt};
          \addlegendentry{MDNet(86.5)}
          \addplot table[x={THRESHOLD},y={VITAL}]{data/otb-precision.txt};
          \addlegendentry{VITAL(88.1)}
        \end{axis}
      \end{tikzpicture}
    }
  \end{subfigure}
  \begin{subfigure}[t]{0.325\textwidth}
    \resizebox{\columnwidth}{!}{
      \begin{tikzpicture}
        \begin{axis}[
            grid=both,
            grid style={line width=.1pt, draw=gray!40},
            ylabel= Precision,
            xlabel= Normalized Location Error Threshold,
            legend pos = south east,
            legend style={font=\small, fill=white, fill opacity=0.6, draw opacity=1, text opacity=1},
            legend cell align={left},
            cycle list name=default,
          ]
          \addplot table[x={THRESHOLD},y={PP[NL]}]{data/otb-normed.txt};
          \addlegendentry{SNLT(80.4)}
          \addplot table[x={THRESHOLD},y={PP[B]}]{data/otb-normed.txt};
          \addlegendentry{SiamRPN++(79.7)}
          \addplot table[x={THRESHOLD},y={RPN[B]}]{data/otb-normed.txt};
          \addlegendentry{SiamRPN(75.8)}
          \addplot table[x={THRESHOLD},y={PrDiMP}]{data/otb-normed.txt};
          \addlegendentry{PrDiMP(83.0)}
          \addplot table[x={THRESHOLD},y={DiMP}]{data/otb-normed.txt};
          \addlegendentry{DiMP(81.9)}
          \addplot table[x={THRESHOLD},y={ATOM}]{data/otb-normed.txt};
          \addlegendentry{ATOM(82.4)}
          \addplot table[x={THRESHOLD},y={MDNet}]{data/otb-normed.txt};
          \addlegendentry{MDNet(82.8)}
          \addplot table[x={THRESHOLD},y={VITAL}]{data/otb-normed.txt};
          \addlegendentry{VITAL(84.2)}
        \end{axis}
      \end{tikzpicture}
    }
  \end{subfigure}
  \caption{Success, Precision, and Normalized Precision Plots on OTB-99-LANG.}
  \label{fig-otb}
\end{figure*}

\begin{figure*}[h]
  \centering
  \begin{subfigure}[t]{0.325\textwidth}
    \resizebox{\columnwidth}{!}{
      \begin{tikzpicture}
        \begin{axis}[
            grid=both,
            grid style={line width=.1pt, draw=gray!40},
            ylabel= Success Rate,
            xlabel= Overlap Threshold,
            legend pos = south west,
            legend style={font=\small, fill=white, fill opacity=0.6, draw opacity=1, text opacity=1},
            legend cell align={left},
            cycle list name=default,
          ]
          \addplot table[x={THRESHOLD},y={PP[BERT]}]{data/lasot-success-consistent.txt};
          \addlegendentry{SNLT(53.1)}
          \addplot table[x={THRESHOLD},y={PP[B]}]{data/lasot-success-consistent.txt};
          \addlegendentry{SiamRPN++(44.1)}
          \addplot table[x={THRESHOLD},y={RPN[B]}]{data/lasot-success-consistent.txt};
          \addlegendentry{SiamRPN(35.8)}
          \addplot table[x={THRESHOLD},y={PrDiMP}]{data/lasot-success-consistent.txt};
          \addlegendentry{PrDiMP(52.8)}
          \addplot table[x={THRESHOLD},y={DiMP}]{data/lasot-success-consistent.txt};
          \addlegendentry{DiMP(54.4)}
          \addplot table[x={THRESHOLD},y={ATOM}]{data/lasot-success-consistent.txt};
          \addlegendentry{ATOM(47.2)}
          \addplot table[x={THRESHOLD},y={MDNet}]{data/lasot-success-consistent.txt};
          \addlegendentry{MDNet(35.3)}
          \addplot table[x={THRESHOLD},y={VITAL}]{data/lasot-success-consistent.txt};
          \addlegendentry{VITAL(35.4)}
        \end{axis}
      \end{tikzpicture}
    }
  \end{subfigure}
  \begin{subfigure}[t]{0.325\textwidth}
    \resizebox{\columnwidth}{!}{
      \begin{tikzpicture}
        \begin{axis}[
            grid=both,
            grid style={line width=.1pt, draw=gray!40},
            ylabel= Precision,
            xlabel= Location Error Threshold,
            legend pos = south east,
            legend style={font=\small, fill=white, fill opacity=0.6, draw opacity=1, text opacity=1},
            legend cell align={left},
            cycle list name=default,
          ]
          \addplot table[x={THRESHOLD},y={PP[BERT]}]{data/lasot-precision-consistent.txt};
          \addlegendentry{SNLT(57.4)}
          \addplot table[x={THRESHOLD},y={PP[B]}]{data/lasot-precision-consistent.txt};
          \addlegendentry{SiamRPN++(48.0)}
          \addplot table[x={THRESHOLD},y={RPN[B]}]{data/lasot-precision-consistent.txt};
          \addlegendentry{SiamRPN(37.2)}
          \addplot table[x={THRESHOLD},y={PrDiMP}]{data/lasot-precision-consistent.txt};
          \addlegendentry{PrDiMP(56.6)}
          \addplot table[x={THRESHOLD},y={DiMP}]{data/lasot-precision-consistent.txt};
          \addlegendentry{DiMP(57.6)}
          \addplot table[x={THRESHOLD},y={ATOM}]{data/lasot-precision-consistent.txt};
          \addlegendentry{ATOM(47.8)}
          \addplot table[x={THRESHOLD},y={MDNet}]{data/lasot-precision-consistent.txt};
          \addlegendentry{MDNet(34.2)}
          \addplot table[x={THRESHOLD},y={VITAL}]{data/lasot-precision-consistent.txt};
          \addlegendentry{VITAL(34.1)}
        \end{axis}
      \end{tikzpicture}
    }
  \end{subfigure}
  \begin{subfigure}[t]{0.325\textwidth}
    \resizebox{\columnwidth}{!}{
      \begin{tikzpicture}
        \begin{axis}[
            grid=both,
            grid style={line width=.1pt, draw=gray!40},
            ylabel= Precision,
            xlabel= Normalized Location Error Threshold,
            legend pos = south east,
            legend style={font=\small, fill=white, fill opacity=0.6, draw opacity=1, text opacity=1},
            legend cell align={left},
            cycle list name=default,
          ]
          \addplot table[x={THRESHOLD},y={PP[BERT]}]{data/lasot-normed-consistent.txt};
          \addlegendentry{SNLT(62.4)}
          \addplot table[x={THRESHOLD},y={PP[B]}]{data/lasot-normed-consistent.txt};
          \addlegendentry{SiamRPN++(55.0)}
          \addplot table[x={THRESHOLD},y={RPN[B]}]{data/lasot-normed-consistent.txt};
          \addlegendentry{SiamRPN(43.9)}
          \addplot table[x={THRESHOLD},y={PrDiMP}]{data/lasot-normed-consistent.txt};
          \addlegendentry{PrDiMP(62.6)}
          \addplot table[x={THRESHOLD},y={DiMP}]{data/lasot-normed-consistent.txt};
          \addlegendentry{DiMP(64.2)}
          \addplot table[x={THRESHOLD},y={ATOM}]{data/lasot-normed-consistent.txt};
          \addlegendentry{ATOM(54.4)}
          \addplot table[x={THRESHOLD},y={MDNet}]{data/lasot-normed-consistent.txt};
          \addlegendentry{MDNet(40.9)}
          \addplot table[x={THRESHOLD},y={VITAL}]{data/lasot-normed-consistent.txt};
          \addlegendentry{VITAL(42.4)}
        \end{axis}
      \end{tikzpicture}
    }
  \end{subfigure}
  \caption{
    Success, Precision, and Normalized Precision Plots on NL-Consistent
    LaSOT~\cite{feng2020real}. Our SNLT outperforms siamese trackers by a large
    margin as the NL descriptions in the NL-Consistent LaSOT uniquely describe
    the targets.
  }
  \label{fig-nl-lasot}
\end{figure*}
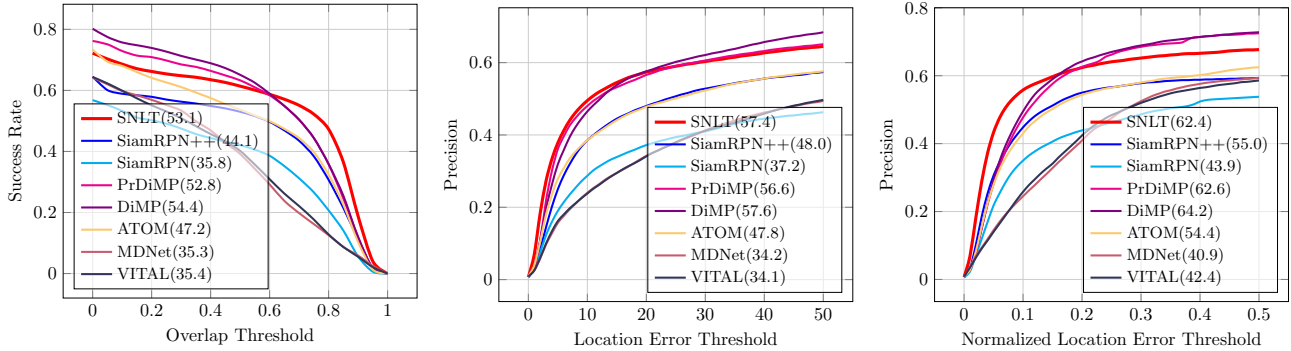

\begin{table}
  \begin{center}
    \begin{tabular}{cccc}
      \hline
      Backbone & NL & Success & Norm. Prec.\\
      \hline
      SiamRPN++ & BERT & \textbf{54.0} & \textbf{63.6} \\
      SiamRPN++ & HGLMM & \textit{50.4} & \textit{60.0} \\
      SiamRPN++ & GLOVE & 49.9 & 59.4 \\
      SiamRPN++ & N/A & 48.9 & 58.0 \\
      \hline
      SiamRPN & BERT & \textbf{46.0} & \textbf{54.6} \\
      SiamRPN & HGLMM & \textit{44.1} & \textit{53.7} \\
      SiamRPN & GLOVE & 43.6 & 52.8 \\
      SiamRPN & N/A & 42.2 & 50.9 \\
      \hline
    \end{tabular}
  \end{center}
  \caption{
    Ablation studies of the proposed SNLT tracker on LaSOT test split. Three
    different sentence embedding models (BERT~\cite{devlin2018bert},
    HGLMM~\cite{burns2019language} and GLOVE~\cite{pennington2014glove}) are
    used to train our SNL-RPN on both SiamRPN~\cite{li2018high} and
    SiamRPN++~\cite{li2019siamrpn++} backbones. The best and second best
    performances are highlighted with bold and italic fonts.
  }
  \label{tbl-ablation-study}
\end{table}

\begin{figure*}[t]
  \begin{subfigure}{\textwidth}
    \centering
    \includegraphics[page=3,trim=0 22cm 0 0,clip,width=\textwidth]{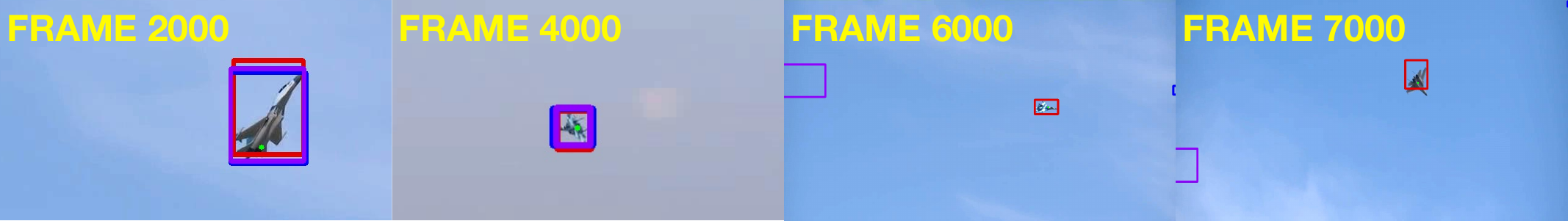}
    \caption{
      The given NL uniquely describes the airplane and helps our tracker (red)
      stay on the target, while SiamFC~\cite{bertinetto2016fully} (purple) and
      SiamRPN++~\cite{li2019siamrpn++} (blue) suffers from model drifts.
    }
  \end{subfigure}
  \begin{subfigure}{\textwidth}
    \centering
    \includegraphics[page=3,trim=0 16cm 0 6cm,clip,width=\textwidth]{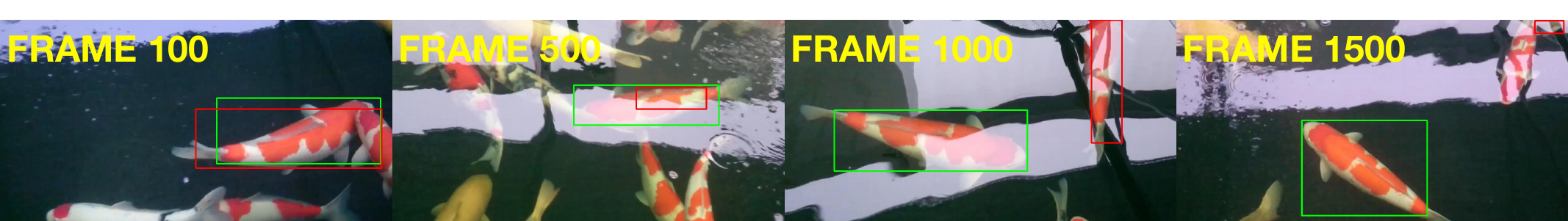}
    \caption{
      The given NL (goldfish swimming among other fishes in the water) does not
      uniquely describes the target. As multiple goldfishes are present in the
      scene, the NL description does not help our tracker (red) to avoid model
      drifting.
    }
  \end{subfigure}
  \caption{The ambiguity of the NL description may affect our tracker.}
  \label{fig-visualization}
\end{figure*}

\begin{figure}
  \centering
  \includegraphics[width=\columnwidth, page=1, trim=1cm 7cm 0 6cm]{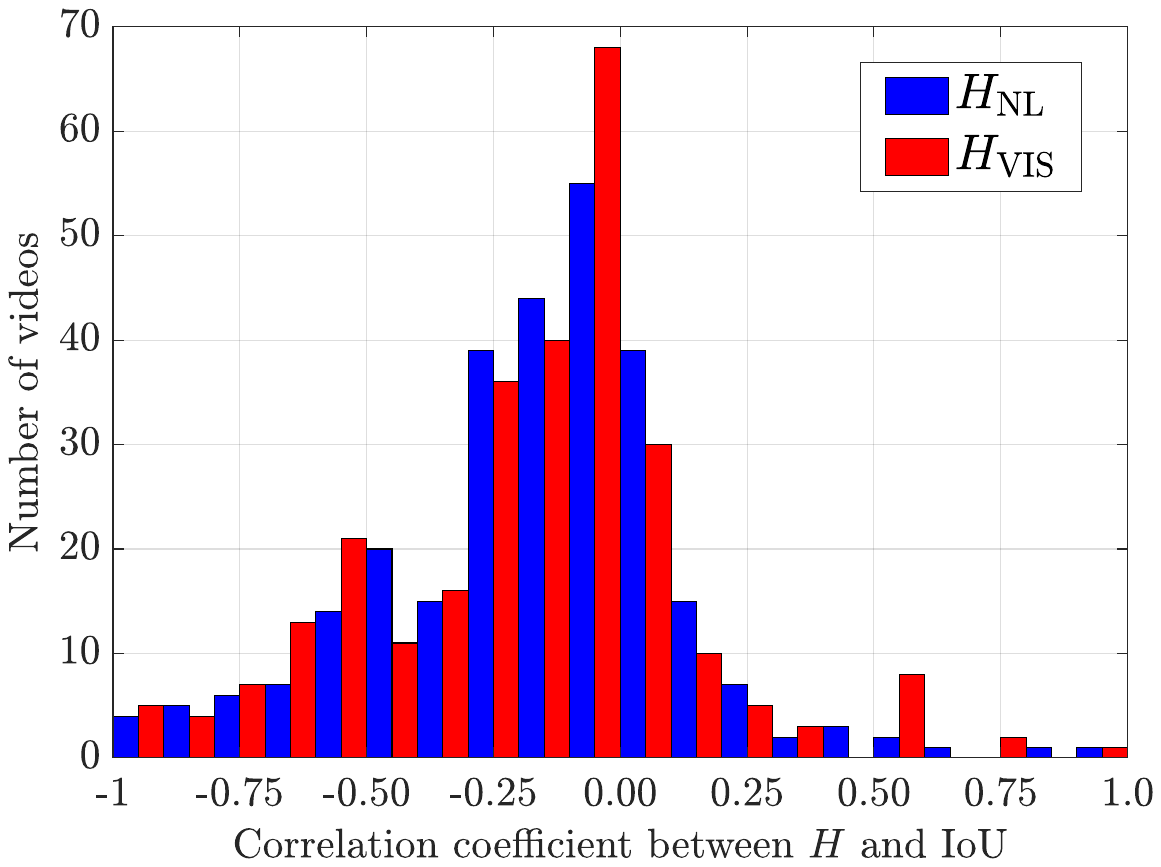}
  \caption{
    Distribution of the correlation coefficient between the entropies defined
    in Eq.~\ref{eq-entropy} and IoU of the predicted and ground truth bounding box.
  }
  \label{fig-corr}
\end{figure}

\begin{table}
    \begin{center}
      \begin{tabular}{ccccc}
        \hline
        &\multicolumn{2}{c}{\textbf{OTB}} &\multicolumn{2}{c}{\textbf{LaSOT}}  \\
        \textbf{Tracker} & Suc. & Norm. & Suc. & Norm.  \\
        \hline
        SNLT & \textbf{0.67} & \textbf{0.80} & \textbf{0.54} & \textbf{0.64}\\
        FENG~\cite{feng2020real} & 0.61 & 0.73 & 0.35 & 0.43\\
        LI~\cite{li2017tracking} & 0.55 & 0.67 & - & -\\
        \hline
        SNLT (SEM) & - & - & 0.40 & 0.48\\
        \hline
      \end{tabular}
      \caption{
        Comparison between LI~\cite{li2017tracking}, FENG~\cite{feng2020real}
        and the SNLT using both bounding box and NL for initialization. ``SEM''
        stands for using the semantic class as the NL description input to the
        SNLT tracker.
        The best performance is highlighted with bold font.
        }
      \label{tbl-lang-performance}
    \end{center}
  \end{table}

\subsection{Comparison with Visual and NL Trackers}
We compare the proposed SNLT with the following state-of-the-art real-time
trackers: SiamRPN++~\cite{li2019siamrpn++}, SiamRPN~\cite{li2018high},
PrDiMP~\cite{danelljan2020probabilistic}, DiMP~\cite{bhat2019learning},
ATOM~\cite{danelljan2019atom}, MDNet~\cite{nam2016learning}, and
VITAL~\cite{song2018vital}. For the fairness of comparisons, we use their
released codes, model weights, and hyper-parameters in all experiments. Success,
Precision, and Normalized Precision Plots on LaSOT~\cite{fan2019lasot} and
OTB-99-LANG test splits are presented in Fig.~\ref{fig-lasot} and
Fig.~\ref{fig-otb} respectively. The SNLT tracker improves the SiamRPN++
baseline on both the LaSOT and the OTB-99-LANG. In OTB-99-LANG, where videos are
typically less than 300 frames, the model drift problem is much less frequent
than that in the LaSOT dataset. Our SNLT improves the performance of the
SiamRPN++ by 0.8 percentage points. In the LaSOT benchmark, as shown in
Fig.~\ref{fig-fps-perf}, the SNLT consistently improves the Siamese trackers by
3 to 7 percentage points, and the top SNLT variant based on SiamRPN++ is very
competitive among state-of-the-art real-time trackers.

Following~\cite{feng2020real}, we evaluate our SNLT on the NL-Consistent LaSOT,
a subset of LaSOT test split in which the NL descriptions uniquely describe the
target in the video (selected by crowd workers).  We show that when the NL
descriptions uniquely describe targets, the proposed SNL-RPN outperforms all
prior works. Success, Precision and Normalized Precision Plots on NL-consistent
LaSOT are presented in Fig.~\ref{fig-nl-lasot}.

We also compare our proposed tracker with best published NL trackers,
Feng~\cite{feng2020real} and Li~\cite{li2017tracking}, on LaSOT and OTB-99-LANG
for the tracking with NL task, \ie tracking with both the bounding
box initialization and the NL description of the target. Evaluations
of~\cite{li2017tracking} on LaSOT are omitted, since no training code was
released for it to guarantee a fair comparison.  As shown in
Tbl.~\ref{tbl-lang-performance}, the SNLT outperforms the best published NL trackers
%on LaSOT and OTB-99-LANG 
by a large margin.

\subsection{Ablation Studies}
\label{sec-ablation}
In this section, we conduct comprehensive experiments and ablation studies
analyzing the performance of our proposed SNLT, SNL-RPN, and the Dynamic
Aggregation Module.

OTB-99-LANG and LaSOT are the only available single object tracking benchmarks
with NL annotations that are publicly available. Only one NL description is
provided for each sequence. Leveraging such NL descriptions to consistently
improve visual tracking ends up very challenging. In
Fig.~\ref{fig-visualization}, we show one case when the NL description would
help our tracker recover from model drift, and another case when the given NL
description does not uniquely describe the target and eventually makes our
tracker drift away to another object that also matches the NL description. This
showcase explains why our SNLT has an even further advantage on the
NL-Consistent LaSOT.

We evaluate the correlation coefficient between the entropies defined in
Eq.~\ref{eq-entropy} and the IoU between the predicted bounding boxes and ground
truth bounding boxes for each video. We plot the distribution of the correlation
coefficient on the LaSOT test split in Fig.~\ref{fig-corr}. The negative
correlation between the entropies and the predicted IoU motivates our Dynamic
Aggregation, in Eq.~\ref{eq-weights}, of the NL Head and the Visual Head in the
SNL-RPN.

\begin{figure}[t]
  \centering
  \includegraphics[width=\columnwidth, page=4, trim=0 17cm 6cm 0]{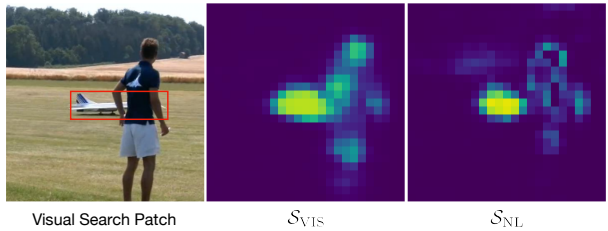}
  \caption{
    Visualization of an example of the Visual Search Patch $X$ (left), Score Map
    from Visual Head $\mathcal{S}_\text{VIS}$ (middle) and Score Map from NL
    Head $\mathcal{S}_\text{NL}$ (right) in the \textit{airplane-13} video of LaSOT. The
    NL description of the target is ``white airplane flying in the air.'' The NL
    Head is more confident and has a more concentrated prediction of the
    airplane than the Visual Head, possibly due to the occlusion of the target.
    Such difference in prediction confidence is reflected in the entropies
    computed by Eq.~\ref{eq-entropy}, then follow Eq.~\ref{eq-weights}, the
    aggregation weights for this example are
    $(w_\text{VIS},w_\text{NL})=(0.39,0.61)$, biasing towards the prediction
    from the NL Head, as expected for this frame. The proposed Dynamic
    Aggregation that gives higher weight to the modality with lower entropy
    results in a more stable tracker.
  }
  \label{fig-heatmap}
\end{figure}

Additionally, an example frame from the \textit{airplane-13} video is visualized
together with the score maps from the NL Nead and the Visual Head in
Fig.~\ref{fig-heatmap}. The proposed Dynamic Aggregation between the two
modalities give higher weights to the modality with lower entropy, resulting in
a more stable tracker.

We test the SNLT with different Language Models, including sentence embedding
model that takes average of word embeddings from GloVe~\cite{he2016identity} and
HGLMM~\cite{burns2019language}, and a BERT sentence embedding
model~\cite{devlin2018bert}. The SNLT were individually trained with the same
procedure as discussed in Sec.~\ref{sec-details}. Comparisons of these trackers
are shown in Tbl.~\ref{tbl-ablation-study}. The proposed SNL-RPN and the Dynamic
Aggregation between vision and language modalities, effectively boost the
tracking performance of both SiamRPN and SiamRPN++.  HGLMM, a more recent
sentence embedding model trained for vision tasks, further pushes the SNLT to a
better performance compared to the GloVe based sentence embedding model. The
sentence embedding models are not fine-tuned for two reasons: 1. They are
pre-trained with large corpuses of texts and the NL tracking datasets are
relatively small. 2. There are several layers after $Z_Q$ that are trained from
scratch. 

The tracking by NL description problem has its unique challenges compared to the
tracking by semantic information~\cite{tripathi2019tracking}. An ablation study
using the ground truth semantic class label, \ie category label in LaSOT (\eg
airplane, person, \etc), as the NL description is reported in
Tbl~\ref{tbl-lang-performance}. Results show that the SNLT learns more than the
semantic class to outperform the baseline SiamRPN++.

Regarding the speed of the proposed SNLT, it only adds a small overhead to
Siamese trackers (most computations from ResNet are shared between Siamese RPN
and SNL-RPN).  As shown in Fig.~\ref{fig-fps-perf}, the SNLTs are around 5\% to
10~\% slower than their corresponding backbones when using the NL-RPN for
inference, still achieving over 50 frames per second.

\section{Conclusion and Future Work}
We present a novel Siamese Natural Language Tracker (SNLT) and Siamese Natural
Language Region Proposal Network (SNL-RPN), which can track a target in a video
given an NL description of the target. With the Dynamic Aggregation Module
between vision and language modalities, our approach enjoys better robustness
than other visual object trackers.  Experiments on challenging datasets
demonstrate that the SNLT outperforms its backbone trackers by a large margin.
Our SNLT and SNL-RPN are generally applicable to all Siamese trackers. Thus, we
expect they will be useful in enhancing future Siamese trackers in pursuit of
state-of-the-art.

\paragraph{Acknowledgment:} This work was partially supported by the US
NSF under Grant No. 1928477.

{
  \small
  \bibliographystyle{ieee}
  \bibliography{bib}
}
\end{document}